\documentclass{article} 
\usepackage{iclr2023_conference_tinypaper,times}

\usepackage{amsmath,amsfonts,bm}

\def\eqref#1{equation~\ref{#1}}

\def\1{\bm{1}}

\DeclareMathAlphabet{\mathsfit}{\encodingdefault}{\sfdefault}{m}{sl}
\SetMathAlphabet{\mathsfit}{bold}{\encodingdefault}{\sfdefault}{bx}{n}

\usepackage{hyperref}
\usepackage{url}
\usepackage{graphicx}
\usepackage{float}
\title{DSF-GAN: DownStream Feedback \\ Generative Adversarial Network}

\author{Oriel Perets \\
Software and Information Systems Engineering\\
Ben Gurion University\\
\And
Nadav Rappoport \\
Software and Information Systems Engineering \\
Ben Gurion University\\
}

\begin{document}

\maketitle

\begin{abstract}
Utility and privacy are two crucial measurements of the quality of synthetic tabular data. While significant advancements have been made in privacy measures, generating synthetic samples with high utility remains challenging. To enhance the utility of synthetic samples, we propose a novel architecture called the DownStream Feedback Generative Adversarial Network (DSF-GAN). This approach incorporates feedback from a downstream prediction model during training to augment the generator's loss function with valuable information. Thus, DSF-GAN utilizes a downstream prediction task to enhance the utility of synthetic samples. To evaluate our method, we tested it using two popular datasets. Our experiments demonstrate improved model performance when training on synthetic samples generated by DSF-GAN, compared to those generated by the same GAN architecture without feedback. The evaluation was conducted on the same validation set comprising real samples. All code and datasets used in this research will be made openly available for ease of reproduction.
 \end{abstract}

\section{Introduction \& Related Work}

\subsection{Introduction}
The use of synthetic tabular data is a topic of high recent interest. Synthetic data serves various purposes, such as preserving privacy by replacing the original samples \citep{yoon_anonymization_2020}, supplementing scarce datasets to develop models requiring more data points and boosting performance \citep{che_boosting_2017}, and balancing datasets to eliminate bias \citep{perets_ensemble_2023}. However, the utility of the generated synthetic data is a critical factor. Data utility measures the effectiveness of synthetic data for a specific task when compared to real samples. One common method for assessing synthetic data utility is by evaluating machine learning efficacy, i.e., comparing the performance of a model trained on synthetic samples with one trained using the real samples, evaluated on the same test set \citep{chin-cheong_generation_2019}. Despite the success of GAN-based approaches in generating realistic and privacy-preserving synthetic samples, the utility of synthetic tabular data lags behind that of real samples in state-of-the-art results \citep{xu_modeling_2019, rajabi_tabfairgan_2022}.

\subsection{Related Work}

A feedback mechanism was first introduced as a way to enhance synthetic samples' similarity. CTAB-GAN \citep{zhao_ctab-gan_2021} and its predecessor CTAB-GAN+ \citep{zhao_ctab-gan_2022} both propose a conditional GAN with a form of feedback that verifies the semantic integrity of synthetic samples' feature values (e.g., a sample whose \texttt{sex=female}, and \texttt{prostate cancer=true} is not semantically correct) \citep{zhao_ctab-gan_2022}. Unlike these approaches, our approach uses the classification or regression performance of a model trained for the specific underlying task the data is aimed at solving. Feedback GAN (FBGAN) \citep{huh_feedback_2019} uses a separate predictor denoted as a function analyzer to optimize the generated gene sequences for desired properties, samples above a threshold are fed back into the discriminator. Similarly, the Feedback Antiviral Peptides GAN \citep{hasegawa_feedback-avpgan_2022} uses AVPs sampled from the generator mid-training to further train the discriminator. While all of the above approaches propose some sort of feedback mechanism, to the best of our knowledge, no documented literature proposed using feedback from the actual classifier or regressor trained for the actual downstream task the synthetic data is generated for.

\section{Methodology}
\label{gen_inst}

The original GAN loss function is given by:
\begin{equation*}
    \min_{G}\max_{D}\mathbb{E}_{x\sim P_{\text{data}}(x)}[\log{D(x;\theta_d)}] +  \mathbb{E}_{z\sim P_{\text{z}}(z)}[1 - \log{D(G(z;\theta_g);\theta_d)}]
\end{equation*}

Where $P_{data}(x)$ is the real data’s distribution. $D(x;\theta_d)$ is the prediction produced by the discriminator $D$, when fed with a mini-batch of real samples $x$, w.r.t the discriminator’s parameters $\theta_d$. $P_z$ is the noise distribution from which the generator $G$ is sampling to generate the synthetic samples, by employing the function $G(z)$ w.r.t the generator’s parameters $\theta_g$. $D(G(z;\theta_g );\theta_d)$ represents the prediction produced by the discriminator $D$ for the a mini-batch of generated sample w.r.t to $\theta_d$. The training procedure consists of two loops optimizing $G$ and $D$ iteratively \citep{goodfellow_generative_2020}.

We demonstrate our architecture using the Conditional Tabular GAN (CTGAN) \citep{xu_modeling_2019} where the loss function is defined as:
\begin{equation*}
\min_{G} \max_{w \in W} V(G,f)=\mathbb{E}_{x \sim P_{data}(x)} [f(x;w)] + \mathbb{E}_{z \sim P_z} [f(G(z;\theta_g);w)] + H
\end{equation*}
Where the discriminator $D$ is replaced with a critic $f$ as proposed by \citep{arjovsky_wasserstein_2017} and $H$ is the binary cross entropy loss between the conditional vector composed on the real samples and the one composed on the generated samples.
Let $loss_G$ denote generator loss, and $loss_D$ denote discriminator loss, our proposed approach (DSF-GAN) facilitates feedback from a downstream task mid-training, adding a loss term of the downstream classifier of regressor to the $loss_G$ scaled by \(\lambda \in \mathbb{R}^+\) to control the magnitude of the feedback term. 

\textbf{Case 1 -- Classification:} let $h(\hat{x}, \theta_h)$ denote a logistic regression model, with loss function given as log-loss:
\begin{equation*}
    L(y, \hat{y}) = -\left( y \log(\hat{y}) + (1 - y) \log(1 - \hat{y}) \right)
\end{equation*}

\textbf{Case 2 -- Regression:} let $h(\hat{x}, \theta_h)$ denote a linear regression model, with loss function given as RMSE:
\begin{equation*}
    L(\theta) = \sqrt{\frac{1}{m} \sum_{i=1}^{m} (h_\theta(x^{(i)}) - y^{(i)})^2}
\end{equation*}

Hence, with the regression or classification loss denoted as $L_f$, the new $Loss_G$ with feedback is:
\begin{equation*}
L_G = \mathbb{E}_{z \sim P_z} [f(G(z;\theta_g);w)] + H + (\lambda*L_f)
\end{equation*}

\section{Experiments}
\label{headings}

We used the base GAN architecture presented in \citep{xu_modeling_2019} for this experiment. For empirically evaluating the increase in synthetic data utility contributed by the downstream feedback, we used two distinct datasets. We trained a DSF-GAN for $\frac{N}{2}$ epochs using the original loss function. For each iteration of training in the remaining epochs ($\frac{N}{2}+1$,..,$N$), we sampled synthetic samples $\hat{x}$ from $G$ and used it to train a logistic regression or linear regression model. We then added the respective loss term to $loss_G$. Post-training, we sampled $n$ samples from the trained model, and used it as a training set for a regression or classification model, we evaluated the model performance using a set-aside validation set comprised of real samples which were excluded from the GAN training. Our results show an increase in utility for the classification and regression tasks and are presented in Tables \ref{results_cla} and \ref{results_reg}. Epochs, batch sizes, datasets, and more details are presented in Appendix A.

\section{Conclusions}
\label{others}

In this work, we propose a novel GAN architecture, with a feedback mechanism from a downstream task (i.e., the original task the dataset was used to solve). This was done by utilizing the loss function of the downstream task trained on the synthetic samples mid-training, and evaluated on the real samples. The empirical experiments show increased synthetic data utility, hence proving the potential of this architecture. Many directions for future work are possible. For example, using other forms of feedback or feedback models. This research is another stepping stone in enabling synthetic data's safe and efficient use in machine-learning tasks.

\subsubsection*{URM Statement}
We acknowledge that all authors in this paper meet the URM criteria of ICLR 2024 Tiny Papers Track.

\bibliography{iclr2023_conference_tinypaper}
\bibliographystyle{iclr2023_conference_tinypaper}

\appendix
\section{Appendix}
\subsection{Figures}
\begin{figure}[h]
\begin{center}
\includegraphics[width=0.8\linewidth]{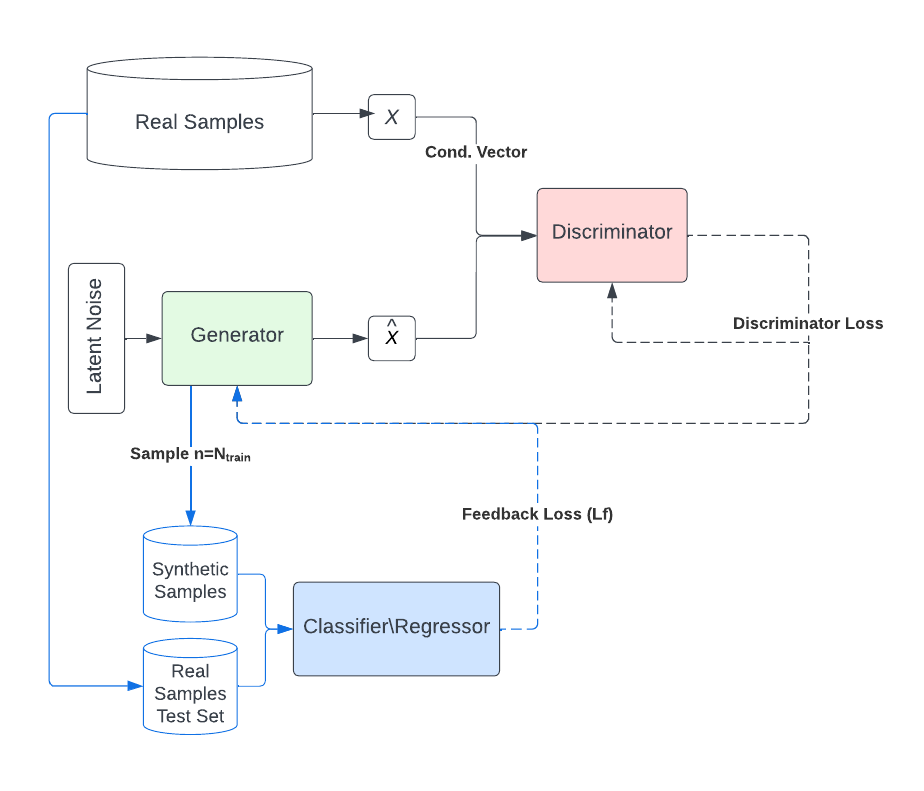}\end{center}
\caption{Downstream Feedback Generative Adversarial Network Schematic Flow}
\end{figure}
\subsection{Experimental Details}
Before training the GAN model, a train, validation split is performed. The training samples are used to train the GAN models, while the validation set, comprised of real samples, is set aside for evaluating the synthetic data which is ultimately generated by the trained models. 
Both the GAN models (base architecture and with the feedback mechanism) are trained separately, using the exact same training set, the same number of epochs, batch size, and other parameters e.g., embedding dimensions and learning rate. \\
To further ensure the stability of the results, we used five-fold Cross-validation and computed the confidence interval, both in the initial split phase before GAN training and in the synthetic data sampling where we sample from the trained model to train the downstream classifier or regressor to evaluate the synthetic data utility.
\subsubsection{Datasets}
\begin{enumerate}
    \item House Price - \hyperlink{https://www.kaggle.com/datasets/shree1992/housedata}{https://www.kaggle.com/datasets/shree1992/housedata}
    \item Adult Census Income - \hyperlink{https://archive.ics.uci.edu/dataset/2/adult}{https://archive.ics.uci.edu/dataset/2/adult}
\end{enumerate}
\subsubsection{Model Training}
\begin{table}[H]
\centering
\begin{tabular}{lccccc}
Data set & n & Epochs & Batch size & $\lambda$ \\
Adult  & 32,560 & 100 & 500 & 1\\
House  & 4,551 & 500 & 200 & 1\\
\end{tabular}
\caption{Training specifics for DSF-GAN}
\label{expirement-times}

\end{table}
\subsection{Results}
\begin{table}[H]
\centering
\begin{tabular}{lcccc}
 & \multicolumn{2}{c}{\textbf{No Feedback}} & \multicolumn{2}{c}{\textbf{Feedback}} \\
Data set & Precision & Recall & Precision & Recall \\
Adult & $0.575\pm0.003$ & $0.441\pm0.007$ & $0.598\pm0.003$ & $0.485\pm0.006$ \\
\end{tabular}
\caption{Results for classification feedback model, precision and recall for a model evaluated on synthetic generated by the base GAN model, and precision, recall for a model trained on synthetic data generated by the DSF-GAN with the feedback mechanism. All models are evaluated using a validation set comprised of real samples}
\label{results_cla}
\end{table}

\begin{table}[H]
\centering
\begin{tabular}{lcccc}
 & \multicolumn{2}{c}{\textbf{No Feedback}} & \multicolumn{2}{c}{\textbf{Feedback}} \\
Data set & RMSE & $R^2$ & RMSE& $R^2$ \\
House & $0.0118\pm1.7e^{-4}$ & $0.3607\pm0.018$ & $0.0115\pm5.8e^{-5}$ & $0.3903\pm0.006$ \\
\end{tabular}
\caption{Results for regression feedback model, RMSE and $R^2$ for a model evaluated on synthetic generated by the base GAN model, and RMSE, $R^2$ for a model trained on synthetic data generated by the DSF-GAN with the feedback mechanism. All models are evaluated using a validation set comprised of real samples}
\label{results_reg}
\end{table}
\end{document}